\documentclass[%
 reprint,
 amsmath,amssymb,
 aps,
]{revtex4-1}
\usepackage{mathtools}
\usepackage{multirow}
\usepackage[export]{adjustbox}

\usepackage{graphicx}
\usepackage{dcolumn}
\usepackage{bm}
\usepackage{hyperref}
\usepackage[english]{babel}
\usepackage[normalem]{ulem}
\usepackage{comment}
\usepackage[draft]{todonotes}   
\usepackage{xcolor}
\usepackage{subcaption} 
\usepackage{booktabs} 
\DeclareMathOperator*{\argmin}{arg\,min}

\begin{document}

\preprint{APS/123-QED}

\title{Informative Neural Ensemble Kalman  Learning}

\author{Margaret Trautner}
\affiliation{Department of Mathematics}
\author{Gabriel Margolis}
\affiliation{Department of Aeronautics and Astronautics}
\author{Sai Ravela}
 \email{ravela@mit.edu}
\affiliation{Department of Earth, Atmospheric, and Planetary Sciences\\ \\ Earth Signals and Systems Group, Massachusetts Institute of Technology, Cambridge, MA 02139, USA}

\date{\today}
\begin{abstract}
 In stochastic systems, {informative approaches} select key measurement or decision variables that maximize information gain to enhance the efficacy of model-related inferences. Neural Learning also embodies stochastic dynamics, but informative Learning is less developed. Here, we propose {\em Informative Ensemble Kalman Learning}, which replaces backpropagation with an adaptive Ensemble Kalman Filter to quantify uncertainty and enables maximizing information gain during Learning. After demonstrating Ensemble Kalman Learning's competitive performance on standard datasets, we apply the informative approach to neural structure learning. In particular, we show that when trained from the Lorenz-63 system's simulations, the efficaciously learned structure recovers the dynamical equations. To the best of our knowledge, Informative Ensemble Kalman Learning is new. Results suggest that this approach to optimized Learning is promising.  
\end{abstract}
 \maketitle    
 \section{Introduction}
\begin{figure*}[htb!]
     \centering
    \includegraphics[width=7in]{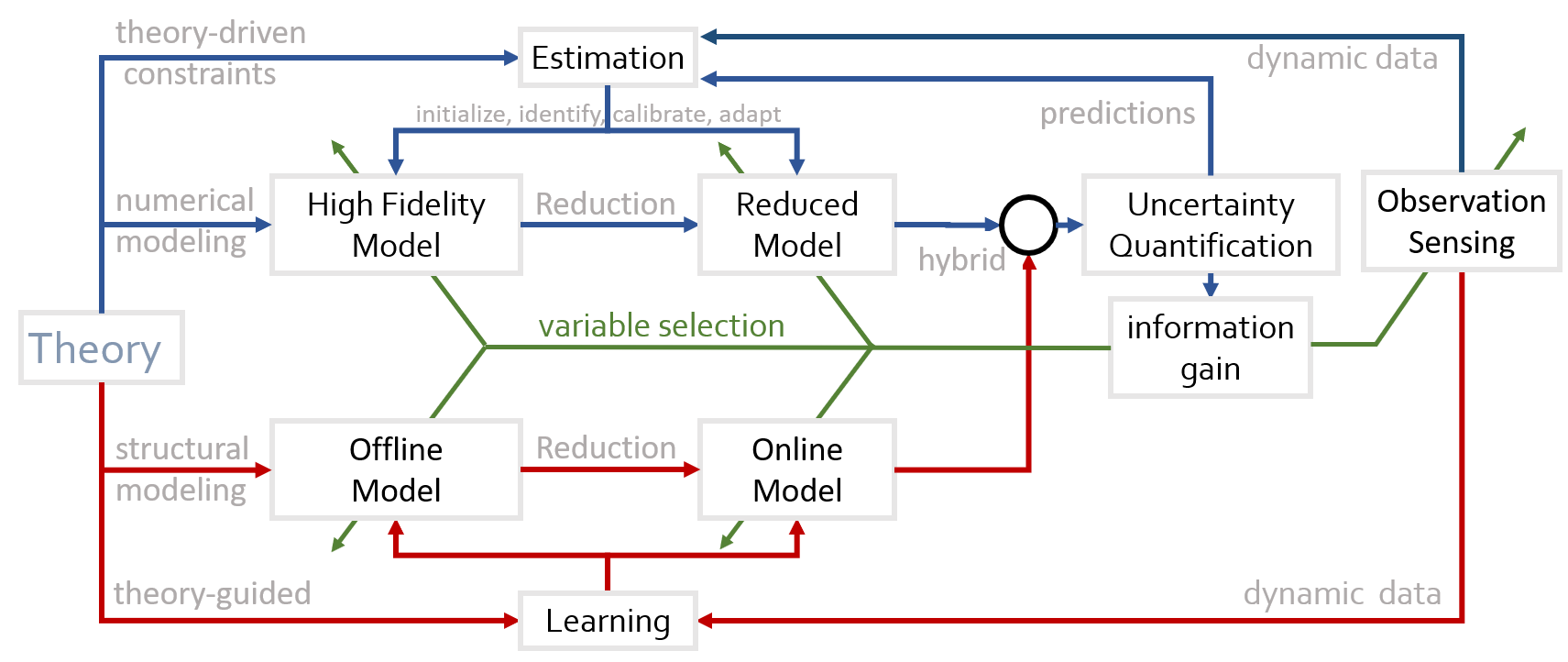}
    \caption{\it The dynamic data-driven paradigm adapts the classical SDO framework (blue paths) with informative approaches (green paths) to improve efficacy, which dynamically select sensors, model parameters, states, and structure to maximize information gain. Learning is playing an increasing role in SDO (red paths). However, informative approaches to Machine Learning are not well developed. Here, we develop Informative Ensemble Kalman Learning with a specific application to learning physics from data and demonstrate general application. }
     \label{fig:dddasdo}
 \end{figure*}

The use of data to dynamically control an executing model and, conversely, using the model to control the instrumentation process is a central tenet of Dynamic Data Driven Applications Systems~\cite{dddasbook} (DDDAS). Applications such as Cooperative Autonomous Observing Systems (CAOS) embody this paradigm in a System Dynamics and Optimization (SDO) loop~\cite{ravela18}. As shown in Figure~\ref{fig:dddasdo}, classical SDO (blue paths) involves model reduction, uncertainty quantification, and estimation. Informative approaches to SDO  dynamically select observations, sensors, model parameters, states, and structures to maximize information gain (green paths). In nonlinear, high-dimensional stochastic processes or systems with epistemic uncertainties, doing so can improve the efficacy of prediction and discovery~\cite{dddasbook}. 

 Learning machines are becoming increasingly popular as standalone and hybrid dynamical systems~\cite{gil19isgeo} to reduce model error, improve uncertainty quantification, and learning dynamics from data. However, the effect of embedding learning in SDO  (red paths) is poorly understood. We do not understand, for example, how externally-trained machines affect closed-loop stability.  We do not have tractable recursive mechanisms for joint retraining and estimation.   
 
 A part of the difficulty stems from poorly developed controls for the learning process itself. The {\it dynamics of Learning}  is often a stochastic, nonlinear, and high dimensional process. However, such a characterization is rare, and rarer is its use to improve the efficacy of Learning. Could SDO be applied to Learning? Would informative approaches enable better or faster Learning? Arguably, a positive outcome will not just advance hybrid systems design but machine learning in general. 
 
 This paper uses neural learning dynamics to develop an informative SDO approach, demonstrates it on standard test problems, and applies it to learn Neural Dynamical Systems~\cite{neuode,trautner19}. Our approach exploits the fact that neural Learning, a parameter estimation problem, entails stochastic dynamics (e.g., due to mini-batches) whose Fokker-Planck equations describe the evolution of network parameter uncertainty~\cite{risken1996fokker}.
 Much like in stochastic nonlinear dynamics, a Monte Carlo approximation to the Fokker-Planck equations leads to tractable Learning. In particular, the Ensemble Kalman Filter~\cite{Evensen03} can train a neural network adjoint-free, exploiting parallelism, and quantifying uncertainty. The proposed adaptive form of the Ensemble Kalman Learner competes well with  backpropagation~\cite{RumelhartHintonWIlliams1986} when evaluated on standard datasets. 

The uncertainty quantification benefit of ensemble methods enables the optimization of Learning. In particular, we maximize information gain between the evolving training error and principal variables of interest and define {\it Informative Learning} as the control of stochastic learning dynamics by maximizing information gain. Informative Learning is sparsity promoting in addition to incorporating uncertainty. It offers a unified approach to several paradigms, including Model Selection, Relevance and Active Learning, and Structure Learning. 

In this paper, we apply Informative Learning to neural Structure Learning (SL) to jointly estimate the optimal structure of a neural network and its parameters from data. In particular, we are interested in SL structure for Neural Dynamical Systems (NDS)~\cite{trautner19}; dynamical systems described in part by Neural Networks. Learning compact NDS from data is a problem of enormous interest, with many applications. However, in addition to generalization and extrapolation issues, it is challenging to gauge interpretability and structural optimality. To overcome these limitations, we use  PolyNet NDS~\cite{trautner19}, which are precisely equal to discrete or continuous dynamical systems with polynomial nonlinearities. Thus, SL for a PolyNet NDS is equivalent to learning dynamical equations from data. The equations themselves are interpretable.  

We apply Informative Learning to learn PolyNet NDS using numerical simulations of the Lorenz-63 system~\cite{Lorenz1963} as training data. We show that the Informative Learning rapidly learns the Lorenz-63 equations to numerical accuracy, {\em ab initio}, from the simulated data, which is an exciting result.

  The rest of this paper is as follows. Section~\ref{sec:rw} describes related work. Section~\ref{sec:nusydo} describes Learning dynamics, section~\ref{sec:enkl} develops Ensemble Kalman Learning, and Section~\ref{sec:ieksl} uses it for Informative Structure Learning.

\section{Related Work}
\label{sec:rw}
There are connections between Informative Learning and Active Learning~\cite{tran19}. However, the latter does not embody a stochastic dynamical perspective. Informative Learning is related to Informative Sensing~\cite{infsense}, Informative Planning~\cite{infpath} and Estimation~\cite{dddasbook}. There is also an application of this concept to Learning~\cite{inflearn} in a manner unrelated to this work. 

The application of the Ensemble Kalman Filter~\cite{Evensen03,Ravela2007} to Learning has received scant interest so far. However, adaptive Ensemble Kalman Learning proposed here offers competitive performance, and Informative Ensemble Kalman Learning is new.  Ensemble Kalman Learning is itself related to Bayesian Deep Learning~\cite{wang2016towards}, but it remains distinct from the extant methodology that typically emphasizes variational Bayesian approaches or Bayesian Active Learning~\cite{tran19}. The ensemble approach only characterizes the network parameter uncertainties as Gaussian even though the network is nonlinear. Further connection of the proposed methodology to sequential Bayesian parameter estimation via Particle Filters and Smoothers~\cite{Arulampalam02atutorial} is natural but left for future work. 

The stochastic dynamics of Learning naturally form a Markov chain ~\cite{ziv17}. Stochastic gradient descent using Kalman-SGD~\cite{patel15} and  Langevin dynamics~\cite{Welling2011BayesianLV} have both been developed, but these are unrelated to our work. Finally, learning physics from data~\cite{raissi2017physicsI} is receiving some attention, but our proposed approach is new.  Structure learning is well developed in Graphical Models~\cite{hastie01statisticallearning}. Neural structure optimization has received some attention~\cite{qianyu}, but our approach still appears to be novel. Note that the presented Informative Ensemble Kalman Learning paradigm is applicable wherever backpropagation is. It can be broadly applied to other learning systems as well.

\section{Neural Systems Dynamics and Optimization}
\label{sec:nusydo}
In this section, we introduce systems definitions for networks, and describe stochastic learning dynamics and informative learning. Let us define a standard Neural Network as  a $N$-stage process~\cite{bryson}:
\begin{eqnarray}
  x_{l+1} &=& F_{l}(x_l,u_l; \alpha_l),  \label{eq:nnproc} \;\;\;0\leq l<N,\\
    y_{N} &=& x_N+\nu_N, \;\label{eq:nnobs}
\end{eqnarray}
where $x_l \in \mathbb{R}^{n_{x,l}}$ are the layer $l$ nodes, $F_l$ is the function, $\alpha_l \in \mathbb{R}^{n_{\alpha,l}}$ are the weights and biases,  $u_l\in \mathbb{R}^{n_{u,l}}$ represents feed-forward\footnote{Feed-forward is used differently in Systems than Learning. Here, the term implies a forward layer-skipping  connection and not Feed-Forward Networks.} and  (e.g. ResNet) or feedback terms (e.g. recurrent network).    The vector $y_N\in \mathbb{R}^{n_{y}}$ refers to (imperfect) training outputs with additive noise $\nu_N\in \mathbb{R}^{n_{y}}$. All the subscripted variables $n_{(\cdot)}$ are positive integers. We may also refer to the network as a single function embedding all layers: 
\begin{equation}    
 x_{N} =F_{NN}(x_1,u; \alpha),\label{eq:stdnn}
\end{equation}
where $\alpha$ is the parameter vector, that is collection of network weights and biases. Training sample denoted ($[x_1, y_N]_s$) is indexed by sample $s$. 

We are also interested in Neural Dynamical Systems,  which are dynamical systems described at least in part by neural networks~\cite{trautner19}. A special  case is a discrete-time autonomous system:
\begin{eqnarray}
  x_{i+1} &=& F(x_i,u_i;\alpha),\label{eq:nds}  \\
    y_i &=& h(x_i)+\nu_i.\label{eq:ndsobs}
\end{eqnarray}
 The network input $x_{i}$ is at discrete time step $i$, $h$ is the ``label" operator~\cite{bryson} and $\nu_i\sim\mathcal{N}(0,R_i)$ is the modeled  additive ``label" noise. These equations also have standard systems interpretations in terms of state, parameter,  control input and measurement/output vectors and operators. 

The network type definitions are necessary for this paper, and for appreciating systems concepts application to Learning. However, they are also incomplete (e.g., missing stochastic neural dynamical system).

\subsection{Two-point Boundary Value Problems}
 
Training Neural Networks using backpropagation is standard~\cite{RumelhartHintonWIlliams1986}. The celebrated backpropagation  algorithm~\cite{RumelhartHintonWIlliams1986}, however, only restates a much earlier solution to multistage two-point boundary value problems~\cite{bryson}(2BVP) commonly used in many areas in engineering and science~\cite{wunschbook,plessixfwi,fdvar}. 

To see this, consider, for example, for a Feed Forward Network defined by Equation~\ref{eq:nnproc} but without control inputs $u$. Training this network in the classical sense using least squares to mimimize expected loss\footnote{Other loss functions are also admissible but not treated in this paper.} with $S$ data samples in the batch, i.e.:
\begin{equation}
\mathcal{L} = \frac{1}{S}\sum_{s=1}^S J(\cdot\;\; ;\;[x_1,y_N]_s), 
\end{equation} is equivalent to the solution to the two point boundary value problem:
\begin{eqnarray}
J(\cdot ; [x_1,y_N])&:=&  \frac{1}{2}(y_{N} - x_{N})^T {R_N}^{-1}(y_{N} - x_{N})\nonumber\\
&&+\sum_{l=1}^N\gamma_l^T \left\{x_{l} - F_{l-1} (x_{l-1};\alpha_{l-1})\right\}.
\end{eqnarray}
The 2-norm uses a Gaussian training error model $\nu_N\sim \mathcal{N}(0,R_N)$. Denote $\gamma\in \mathbb{R}^{n_{l,x}}$ as the Lagrange multiplier. Backpropagation emerges from the normal equations~\cite{bryson}:
\begin{eqnarray}
\texttt{Input:}&& x_1:=x_{1,s}, \\
\texttt{Forward:} && 0<l\leq N\nonumber\\
x_{l,s}&=& F_{l-1}, (x_{l-1,s};\alpha_{l-1})\;\;\\
\texttt{Terminal Error:}&&y_N:=y_{N,s}\\
\gamma_N &=& R^{-1}(y_{N} - x_{N}), \label{bvp2}\\
\texttt{Backward:} && N>k>0\nonumber\\
\gamma_k &=& \nabla_{x_k} F_{k}^T \;\gamma_{k+1}\\
\texttt{Parameter Gradient:} && 0\leq j <N\nonumber\\
\frac{\partial{J_s}}{\partial{\alpha_j}} &=&[\nabla_{\alpha_j} F_j]^T \gamma_{j+1}.\label{eq:parmgrad} \;\;
\end{eqnarray}
The term $\nabla_{\alpha_j} F_j$ is the parameter Jacobian of $F_j$ for parameter vector $\alpha_j$ at operating point $x_j$. Iterative updates typically follow from Equation~\ref{eq:parmgrad}. For example, gradient descent\footnote{Other iterative optimization schemes are also amenable to the methodology in this work but not discussed.} with some learning rate $\tau$ is:
\begin{equation}
\Delta \alpha_l = -\tau \frac{1}{S} \sum_{s=1}^{S} \frac{\partial{J\left(\alpha_l;\; [x_1,y_N]_s\right)}}{\partial{\alpha_l}}.\label{learndyn}
\end{equation}

Similar equations are easily derived for recurrent systems described by Equations~\ref{eq:nnproc}-\ref{eq:nnobs} using a receding horizon/rollout~\cite{bryson}. 2BVP is commonly applied to  dynamical system shown in Equations~\ref{eq:nds}-\ref{eq:ndsobs}, thus are directly applicable to NDS. 

Learning is {not} usually described as 2BVP\footnote{See course Machine Learning with Systems Dynamics and Optimization at MIT--http://essg.mit.edu/ml}, but the connection enables firm methodological footing. Consider, for example, the vanishing/exploding gradient problem in backpropagation.  Analogous to the classical Finite-time Lyapunov exponent~\cite{HALLER2001248}, estimating a  {\em Finite-Depth Lyapunov Exponent} backward in layers using the adjoint equation diagnoses  the issue, which may then be used to construct  bypass connections through $u$. The full development of this idea is out of the scope of this paper. 
\subsection{Stochastic Learning Dynamics}
We now turn to the dynamics of learning using Equation~\ref{learndyn}. The popular use of  ``mini-batches" in Learning turns  gradient descent into a stochastic process. To see this, express  the  minibatch average loss-function gradient, defined for minibatch set $B_i$ in iteration $i$ as: 
\begin{equation}
    \overline{\nabla J(\alpha_i)}=\frac{1}{|B_i|}\sum_{s\in B_i}^{} \frac{\partial{J\left(\alpha_i;\; [x_1,y_N]_{s}\right)}}{\partial{\alpha_i}},
\end{equation} 
where, $\alpha_i$ is the entire network's parameter vector at iteration $i$ as represented in Equation~\ref{eq:stdnn}. 

We now model the minibatch expected loss as the batch expected loss  plus a deviation:
\begin{equation}
     -\tau\overline{\nabla J(\alpha_{i})}= \mu(\alpha_{i}) +  w(\alpha_{i}),
     \label{eq:stochproc0}
\end{equation}
where $\mu(\cdot)$ is the iteration-dependent full-data gradient mean function and $w(\cdot)$ the deviation. We further model $w$ as a random field:
\begin{equation}
    w(\alpha_i)= \sigma(\alpha_i)\;\eta_i,
\end{equation}
where, $\eta_i$ is a zero mean unit variance I.I.D. random vector and $\sigma$ is the square-root of the covariance. Such a Gauss-Markov field, for example, can be approximated from an ensemble of training-data minibatches sampled from the batch. 

Equation~\ref{eq:stochproc0}  allows the formulation of a corresponding continuous-time stochastic learning process in the sense of Ito~\cite{risken1996fokker}: 
\begin{equation}
   d\alpha_t = {\mu_t}(\alpha_t)\;dt+{\sigma_t}(\alpha_t)\;\eta_t.
\end{equation}
A Master equation for evolution of the probability density function $p_\alpha(A_t=\alpha,t)$ associated with parameter vector $\alpha$ is the Fokker-Planck equation~\cite{risken1996fokker}, that is: 
\begin{equation}
\frac{\partial p_\alpha}{\partial t} = -\sum_{j=1}^n \frac{\partial}{\partial \alpha_j}\left[\mu_{t,j}\;p_{\alpha}\right]+\sum_{k=1}^n\sum_{l=1}^n \frac{\partial^2}{\partial \alpha_k \partial \alpha_l} \left[D_{t,kl}\;p_\alpha\right],
\end{equation}
where, $D_t = \sigma_t\sigma_t^T$ is the diffusion tensor, and $\mu_t$ is drift. Starting with suitable initial condition of $p_\alpha$ and boundary conditions, the  stochastic dynamics of learning specifies the evolution of the uncertainty in estimates of its parameters. 

\subsection{Informative Learning}
Stochastic learning dynamics enables Informative Learning. The (conditional) mutual information between variables of interest can optimize learning. To see how, lets define a few  random variables of interest:   data $\mathcal{D}$, the minibatch $\mathcal{D}_{s,t}$,  prediction error $\mathcal{E}_t$, parameter random vector $\mathcal{A}_t$, parameter vector subset $\mathcal{A}_{s,t}$, network structure $\mathcal{S}_t$ and sub-structure $\mathcal{S}_{s,t}$ (e.g. sub-network, terms or features). 

Letting $0\leq t < t' \in \mathbb{R}$, we can thus quantify conditional mutual information for different types of information gain~\cite{Cover2006}:
\begin{itemize}
    \item Transfer mutual information: $\mathcal{I}(\mathcal{A}_t:\mathcal{D}|\mathcal{A}_0)$ is useful to quantify the efficacy of learning.
    \item Input Selection:  $\mathcal{I}(\mathcal{D}_{s,t'>t}:\mathcal{E}_t|\mathcal{D})$, which is also applicable to Data/Feature selection and active learning.
    \item Parameter Selection: $\mathcal{I}(\mathcal{A}_{s,t'}:\mathcal{E}_t|\mathcal{A}_t)$
    which enables resource-constrained and other forms of learning. 
    \item Structure Learning: $\mathcal{I}(\mathcal{S}_{s,t'}:\mathcal{E}_t|\mathcal{S}_t)$, to adapt nodes, layers, and activations for optimizing network structure. 
\end{itemize}   
In Section~\ref{sec:ieksl} we apply this approach to structure learning. Unfortunately, solving the Fokker-Planck equation directly to quantify these terms is only feasible for very low dimensional networks. In practice, we must resort to sampling methods as proposed in the next section.  

\section{Ensemble Kalman Learning}
\label{sec:enkl}

Sampling to solve the Fokker-Planck leads to an ensemble approach~\cite{Evensen03,Ravela2007} to Learning, which is akin to its use in parameter estimation for nonlinear dynamics. However, note that using variational and Bayesian estimation perspectives can also derive this approach. The Ensemble Kalman Filter (EnKF)~\cite{Evensen03,Ravela2007} and smoother, leveraging sample approximation to gradients for inference, are popular alternatives to 2BVP. 

For the following initial discussion, we use  Equation~\ref{eq:nds}-\ref{eq:ndsobs} as a discrete dynamical system and model the Gaussian observational noise at step $i$ as isotropic, i.e., $\nu_i \sim \mathcal{N}(0,R_i=r^2 I)$, $r\in \mathbb{R}$. Further, define $X_i=[x_{i,1}\ldots x_{i,E}]$ to be a matrix of $E$ state (column) vector samples   obtained by solving $F$ model equations from  an initial condition ensemble $X_{i-1}$ at the previous time step. Define the observational projection of the ensemble as $Z_i = [h(x_{i,1})\ldots h(x_{i,E}) ]$, and $Y=[y_{i,1}\ldots y_{i,E}]$ as an ensemble of observations\footnote{Perturbed observations are used here for simplicity. This is not strictly necessary.}. We adopt the notation that $\tilde{Q}$ is a deviation matrix obtained by removing the mean column vector of $Q$ from its columns. A number of estimators now can be defined.
\paragraph{State Estimation Filter:} The filter state estimate $X_i^+$ may be written as~\cite{Evensen03}:
\begin{eqnarray}
    X_i^+ &=& X_i + \tilde{X}_i\tilde{Z}_i^T [\tilde{Z}_i\tilde{Z}_i^T + R_i]^{-1} (Y_i - Z_i)\nonumber \\
    &=& X_i M_{x,i}, \label{enkf}
\end{eqnarray}
where, $M_{x,i}$ is obtained by perturbing the state variable alone keeping others, namely the parameters and control inputs, constant. 
Lagged estimates of a past state are obtained as~\cite{Ravela2007}:
\begin{equation}
X_{i-k}^+ = X_{i-k} M_{x,i}\;\; 0\leq k\leq i. \label{eq:lenkf}
\end{equation}
\paragraph{Fixed-Lag (moving window) Smoothers:} Fixed-lag smoothers from observations in a moving window $W>0$ can be built easily~\cite{Ravela2007}:
\begin{eqnarray}
    X_i^{++} &=& X_i^+ \prod_{j=1}^{W} M_{x,i+j}, \label{eq:fixlag}
\end{eqnarray} 
where, update matrices $M_{x,i+j}$ at step $0<j\leq W$ are computed when  measurements are available,  or set to identity otherwise. A constant-time recursive estimator can update states in a fixed-lag smoother as it marches from one discrete time step to the next $i$~\cite{Ravela2007}. 
\paragraph{Fixed-Interval Smoothers:} State estimates over a fixed interval $[0, W]$ is also easily obtained:
\begin{eqnarray}
    X_i^{++} &=& X_i^+ \prod_{j=i+1}^{W} M_{x,j},\; 0\leq i <W.\label{eq:fixint}
\end{eqnarray} 
The filter updates $M_{x,i}$ over interval $[0,W]$ are similarly calculated as the fixed lag case. Smoothing requires exactly one forward filtering pass and one backward recursion~\cite{Ravela2007}. Interval smoothers are comparable in principle to 2BVP.
\paragraph{Ensemble Control:} Akin to the state estimation problem, the ensemble approach can be used for optimal control also.
\begin{eqnarray}
   U^+_{i}& =& U_{i} M_{u,i+k},\;k>0.\label{eq:enscon}
\end{eqnarray}
 Control perturbations at time step $i$ generate $M_{u,i+k}$ from a fixed initial condition and system $F$ parameters. The controller uses {\em goal} ensemble $Y_{i+k}$ at step $i+k$ that is to be satisfied up to a performance index $R$ which controls desired precision. Fixed-interval smoother form akin to Equation~\ref{eq:fixint} can be used for  trajectory/path optimization. Fixed-lag form akin to Equation~\ref{eq:fixlag} implements a receding horizon Ensemble Model predictive control scheme. This control approach is particularly well suited to direct sensor-to-actuator control~\cite{viewpointcontrol} and is alternative to adaptive control~\cite{ankitpx4} and learning control.

\paragraph{Parameter Estimation:} Let  $A_i=[\alpha_{1}\ldots \alpha_{E}]_i$ is the matrix of  parameter samples at time step $i$. The parameters are constant time, so the parameter ensemble persists from one time step to the next. Only when an observation arrives does it update, which is:
\begin{equation}
A_i^+ = A_i M_{\alpha,i+k},\;k\geq 0,\label{eq:parmest}
\end{equation}
where, an initial ensemble of parameters with fixed initial condition and control input sequence is used for a $k-step$ ensemble simulation to derive a parameter update $M_{\alpha,k}$. The matrix $R$ in this case is just the observational covariance.

\begin{figure}[htb!]
    \centering
\includegraphics[width=3.25in]{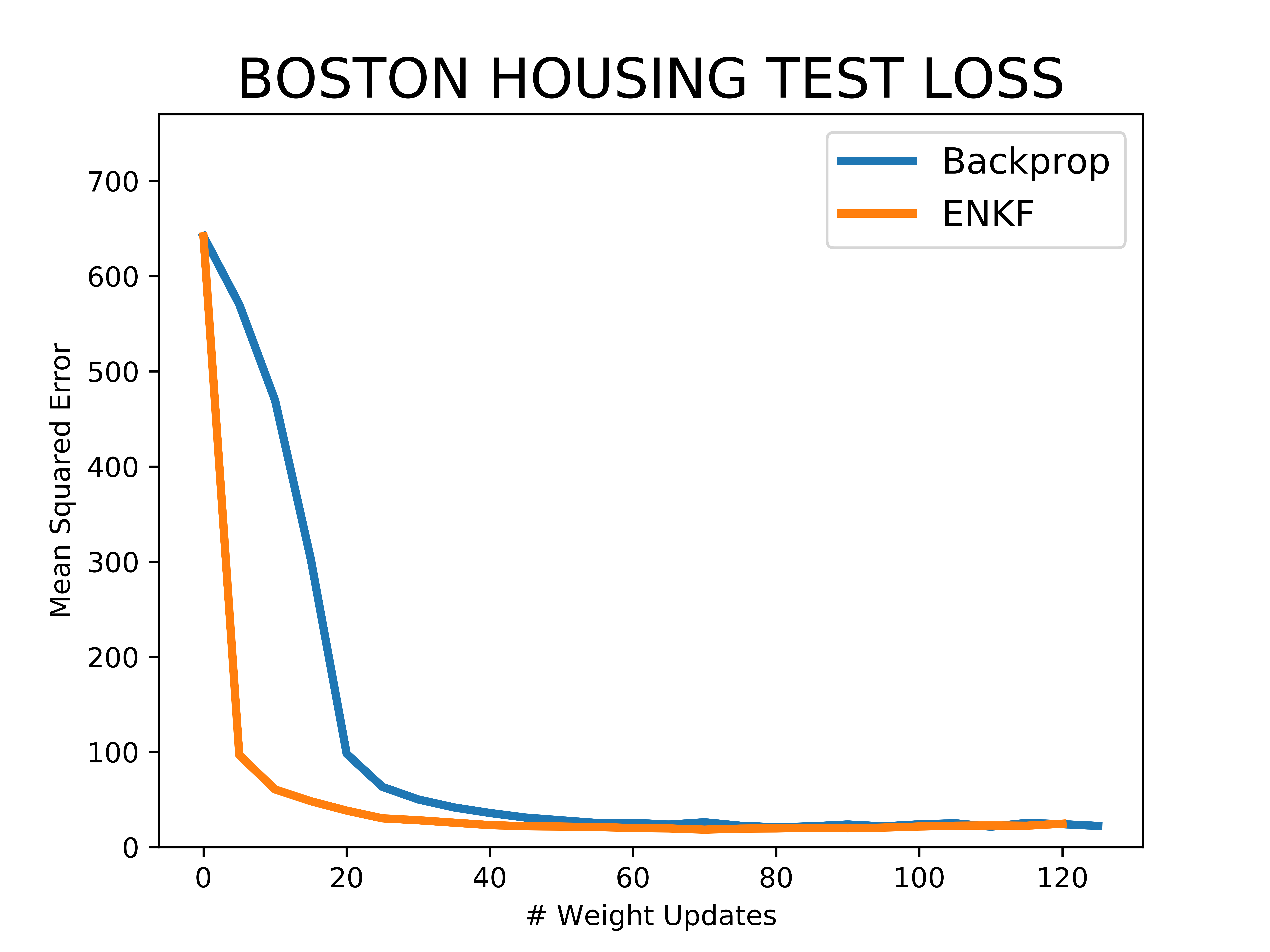}
\includegraphics[width=3.25in]{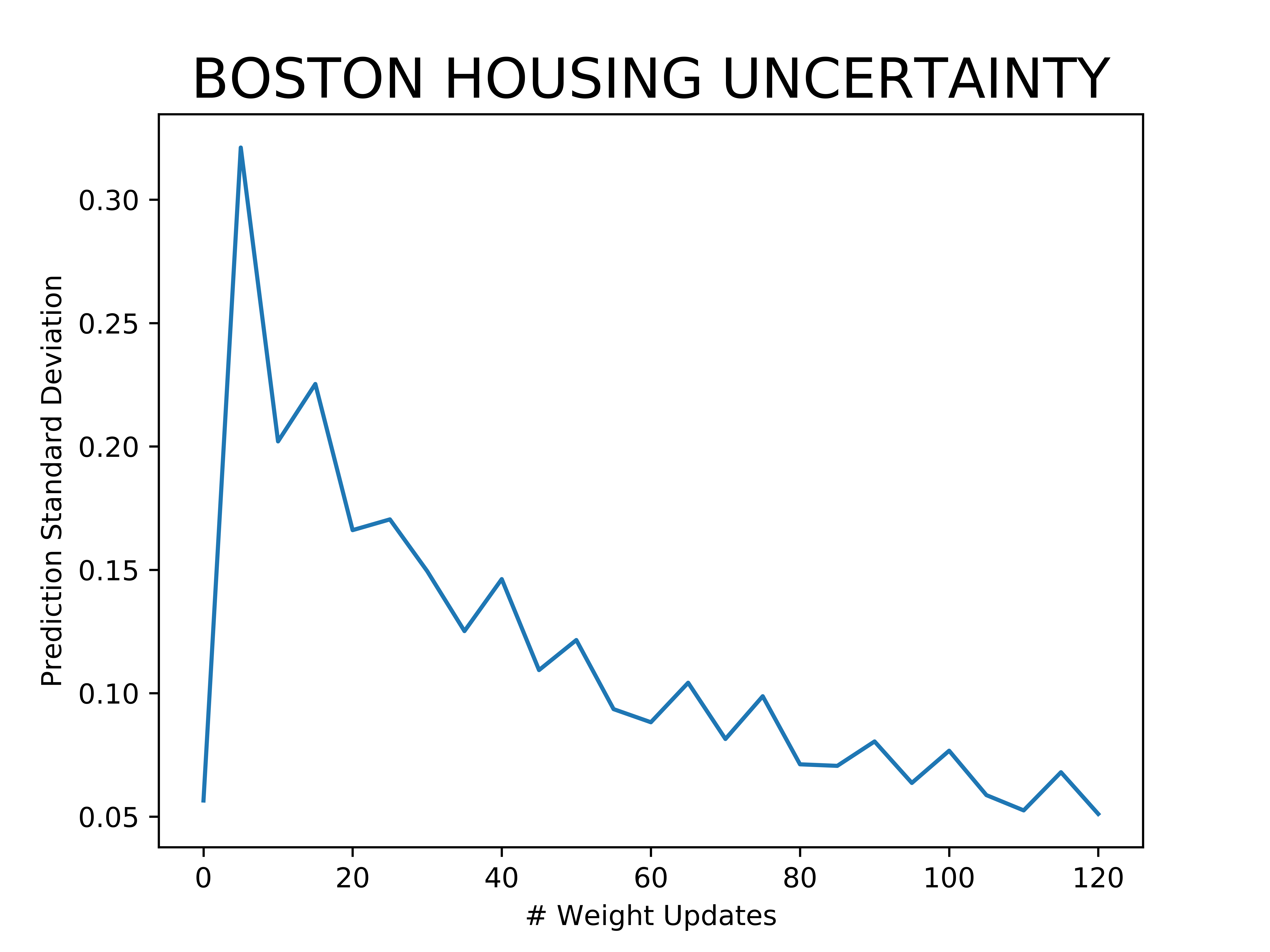}
\caption{\it The Ensemble Kalman Learner offers competitive performance on Boston Housing datasets relative to backpropagation. The top graph shows the test error, and the bottom depicts parameter uncertainty. It is initially small, but the total prediction error is largely due to bias not shown. }
    \label{fig:enkfboston}
\end{figure}

\begin{figure}[htb!]
    \centering
\includegraphics[width=3.25in]{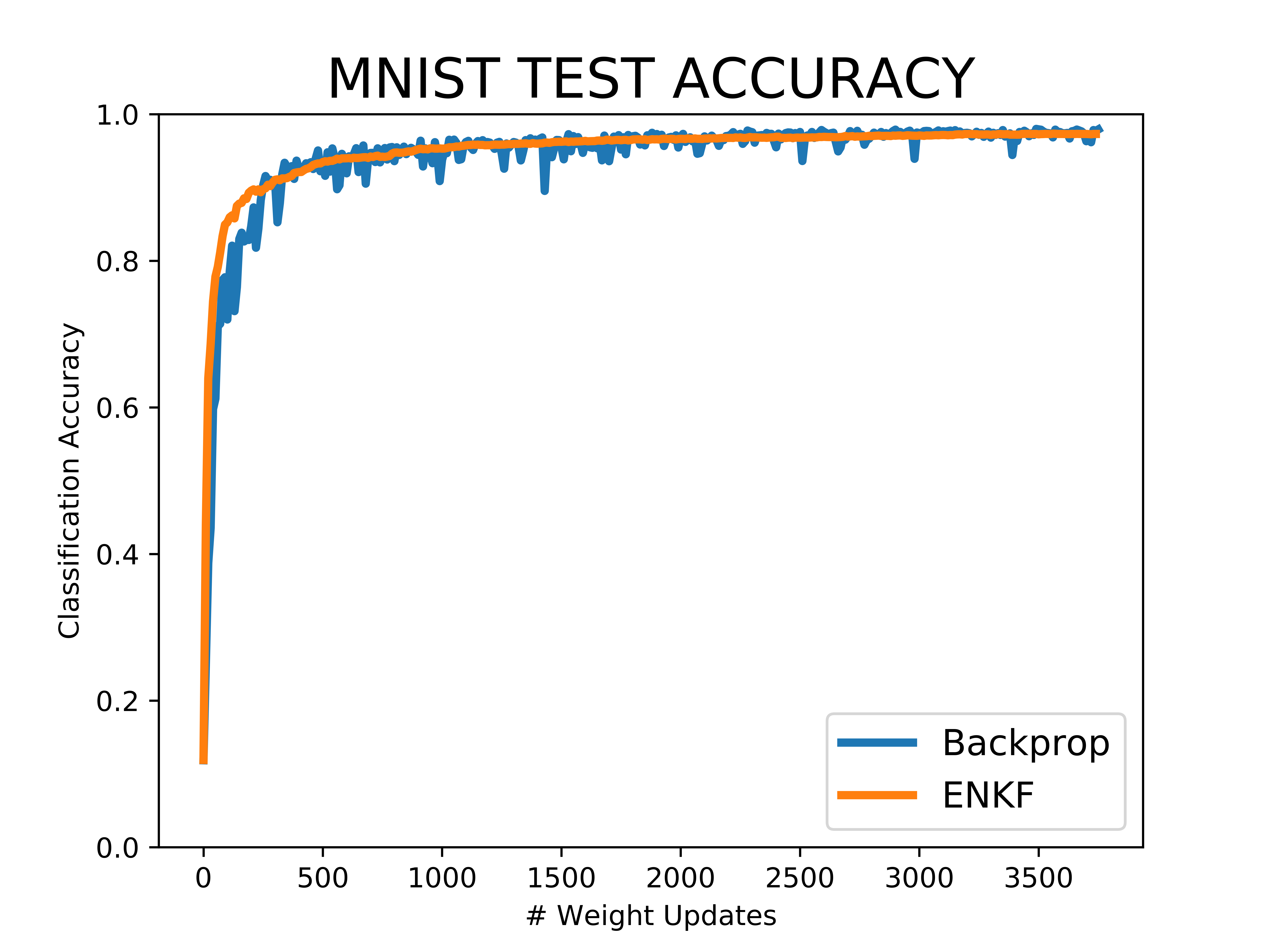}
\includegraphics[width=3.25in]{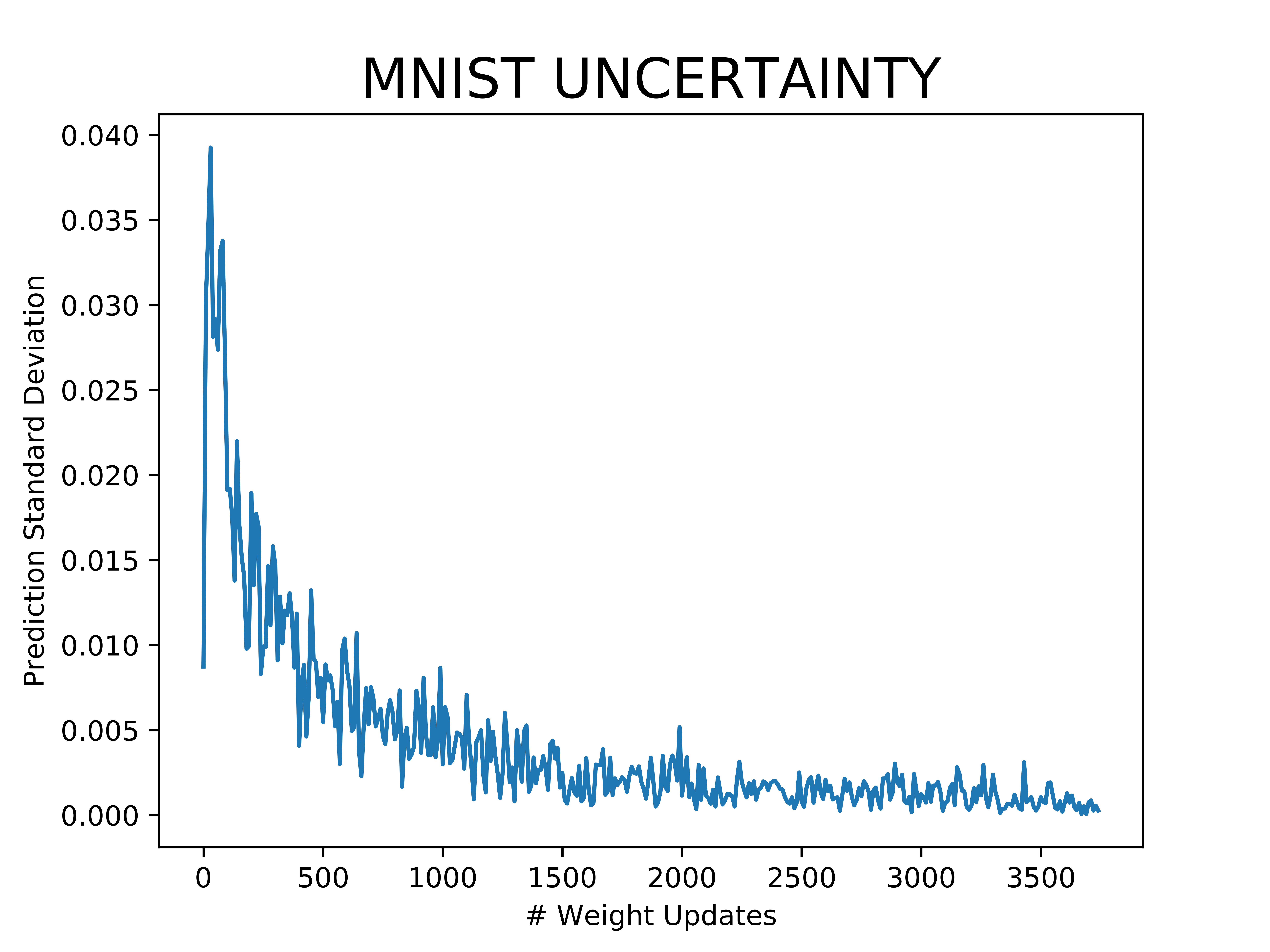}
\caption{\it The Ensemble Kalman Learner offers competitive performance on the MNIST dataset relative to backpropagation. The top graph shows test accuracy, and the bottom depicts parameter uncertainty. It is initially small, but the total prediction error is largely due to bias not shown.}
    \label{fig:enkfmnist}
\end{figure}

\paragraph{Ensemble Learning:} Parameter estimation immediately provides the basis for neural learning and it is best to consider the form of Equation~\ref{eq:stdnn} in this context. Further, for Learning, $i$ is a discrete learning iteration (during gradient descent) using minibatches $B_i$ as previously discussed. In particular, consider $A_1$ to be the initial parameter ensemble (generated with a first-guess Gaussian distribution) and $B_i$ be the minibatch of size $S_i$ at  iteration $i$. Then, the Ensemble Kalman Learning rule is:
\begin{equation}
A_{i+1}= A_i \frac{1}{S_i}\sum_{s\in B_i} M_{\alpha,i,s},\;i>0.\label{eq:enkl}
\end{equation}
In this formulation, $M_{\alpha,i,s}$ is the the update at iteration $i$ using an ensemble simulation of parameter-perturbed neural networks for each  training sample $[x_1,y_N]_{s\in B_i}$.  

Thus, $S_i\times E$ parallel simulations are performed. The ``noise model" ($R$) specifies a tolerance or performance index in achieving training outputs. It can represent training label noise.  The resulting $A_{i+1}$ are then the parameters at iteration $i+1$. The same approach applies to recurrent systems (with rollout), and neural dynamical systems in fixed-lag or fixed-interval or filter forms. This approach strongly contrasts with 2BVP and may be applied to stochastic neural dynamical systems and learning systems in general. 

Ensemble Kalman Learning has several interesting properties. Network linearization and analytical parameter gradients are unnecessary; loss functions are not limited. Parameter uncertainty estimates, directly obtained, further allows quantifying information gain. In contrast to 2BVP, all layer weights update in parallel. Parallel simulations reduce the computational expense, and for small parameter ensembles, the update is compact. We have not constructed learning rules using the fixed-interval, fixed-lag, and ensemble control forms applicable to more complex hybrid neural dynamical systems.

\subsection{Ensemble Kalman Learning Examples}
We conducted examples with the Boston Housing~\cite{bostonhousing} and MNIST~\cite{mnist} datasets\footnote{Obtain code from github.com/sairavela/EnsembleTrain.git}.  For Boston Housing,  we use a neural network with two $32$-neuron hidden layers, ReLU activations, least-squares loss function, minibatch of size $16$, and $100$-member parameter ensemble. An IID zero-mean Gaussian with a standard deviation of $0.01$ provides initial samples, and the target tolerance is $r=0.01$. SGD with a learning rate of $0.1$ controls backpropagation. The results (see Figure~\ref{fig:enkfboston}) show that the Ensemble Kalman Learner achieves a converged error similar to tuned backpropagation within five epochs.

The MNIST dataset~\cite{mnist} network architecture consists of two batch-normalized convolutional layers, max-pooling, and ReLU activations, followed by a single ReLU-activated linear layer of width 10, finally followed by a softmax-activated categorical output layer. We use a least-squares loss function, minibatch size $16$, parameter ensemble size $1000$, and a target error tolerance of $0.015$ to match the observed performance of a highly-performing backpropagation-trained network. Furthermore, the target error tolerance adapts as ensemble variance reduces, up to a lower bound of $0.0015$. SGD with a learning rate of $1.0$ parameterizes backpropagation. The Ensemble Kalman Learner achieves a final test accuracy of $97.1\% $, competing well with backpropagation at  $97.9\% $. It does this while maintaining better stability at a high learning rate, see Figure~\ref{fig:enkfmnist}.

The ensemble approach is simple to implement, quantifies uncertainty, and can transform Learning using distributed computation.  We leverage this to demonstrate efficacious Neural Structure Learning.

\section{Informative Ensemble Kalman Learning}
\label{sec:ieksl}
Ensemble Kalman Learning enables the informative approach discussed in Section~\ref{sec:nusydo}. For notational convenience in this section,  define scalar variables in lower case $e\in \mathbb{R}$, vector variables as $\mathbf{e}\in\mathbb{R}^n, n>1$, ensemble matrices in upper case $E\in \mathbb{R}^{n\times S}$ of $S>0$ samples, and random vectors as $\mathcal{E}=[\mathcal{E}_i]_{i=1}^n,\mathcal{E}_i\in \mathbb{R}$. We define the mutual information  $\Psi$ between two random vectors $\mathcal{A}=[\mathcal{A}_i\in\mathbb{R}]_{i=1}^m$ and $\mathcal{E}=[\mathcal{E}_j\in\mathbb{R}]_{j=1}^n$ as the pairwise mutual information between their  component variables~\cite{Cover2006}. That is, 
\begin{eqnarray}
    I(\mathcal{A}:\mathcal{E}) &:=& \left[ \;\Psi_{i,j}\;\right]_{m\times n}(\mathcal{A}:\mathcal{E}),\\
    \Psi_{i,j}(\mathcal{A}:\mathcal{E}) &:=& -\frac{1}{2} \ln\left(1 -\rho^2(\mathcal{A}_i,\mathcal{E}_j)\right).
\end{eqnarray} 
 The correlation coefficient $\rho$ is empirically estimated from the respective ensembles $A$ and $E$. Note that table entry $\Psi_{ij}\geq 0$ with a numerically bounded maximum in practice, represents association. 
 
 Maximizing information gain by selection entails selecting a sparse subset of elements of $\mathcal{A}$ or pairings of $\mathcal{A}$-$\mathcal{E}$  that maximizes the cumulative mutual information in $I$, which is an NP-hard $\ell_0$ problem. Another popular alternative is $\ell_1$, but $\ell_1$ is a rather weak approximation to $\ell_0$, and greedy solutions to $\ell_0$~\cite{greedyvsl1,Liu_2016L0} can outperform while being fast. In problems with a large number of variables to select from, selection by iterative re-weighting for an eventual pruning suffers from additional dimensionality concerns. 
 
 Here, we adopt a simple greedy approach that is then applied in the next section to Structure Learning. Other applications as described in Section~\ref{sec:nusydo} are not discussed in this paper. Pairwise mutual information in table $I$ can be ranked (in decreasing order) and sorted into a vector variable ${I}^*$, that is:
\begin{eqnarray}
    {I}^* &:=& \left[\Psi^*_k\right]_{k=1}^{mn},\\
    \Psi^*_l&\geq& \Psi^*_{l+1},\;0<l<mn,\\
    \Psi^*_{k}&:=& \Psi_{i_k,j_k},\;1\leq i_k \leq m, 1\leq j_k \leq n. \nonumber
\end{eqnarray} 
 This forms the basis for variable selection to maximize information gain. Here, we use information criteria (akin to AIC or BIC)~\cite{hastie01statisticallearning} to penalize the terms selected in terms of model complexity using function $C(k)$ that monotonically increases with number of terms $k$ selected. That is: 
\begin{eqnarray}
\label{eq:mimax}
k^* &=& \argmin_k\;\underbrace{ \sum_{l=1}^k\left[ 1 - \frac{1}{\Psi^\#}  \Psi^*_l\right] }_{\textrm{Decreasing}}+ \underbrace{C(k)}_{\textrm{Increasing}},\\
\Psi^\# &=&\sum_{o=1}^{mn} \Psi^*_o,\;\;1\leq k \leq mn. \nonumber
\end{eqnarray}
 The ordering of pairwise mutual information allows for a greedy algorithm to optimize Equation~\ref{eq:mimax}. The selected pairings correspond to random variables $\mathcal{A}_1\ldots \mathcal{A}_{k^*}$ from which unique members may be selected, as needed.

An informative Ensemble Kalman Filtering approach jointly solves the parameter estimation and variable selection problem. In particular, Equations~\ref{eq:enkl} and Equation~\ref{eq:mimax} are alternated as we show in the next section in a Structure Learning example. Such alternation has an Expectation Maximization~\cite{emgentle} as well as Gibbs Sampling~\cite{gelfandgibbs} interpretation, but here, the framework is akin to Informative approaches alternation of estimation with information gain-based selection. 

\subsection{Informative Structure Learning}
 Neural Structure Learning is a difficult problem; consider just learning $y=x^2$ with a $tanh$ activation node is hard ($tanh$ has no even Taylor expansion terms)~\cite{inflearnagu}. That is, the neural model is structurally poor and is somewhat uninterpretable. In general, in addition to these problems, lack of  generalization and extrapolation are  confounding factors. The development and assessment of methodology is thus difficult, particularly if one wishes to evaluate the efficacy of the proposed approaches. 
 
 As discussed in the introduction, one way to develop methodology is to test it learn the structure from data generated by a known function. Learning the structure of neural dynamical systems~\cite{trautner19} trained from the non-trivial and large class of polynomial dynamics is one such possibility. 

Recognizing that neural networks with multiplicative gates ~\cite{trautner19} (PolyNet) can represent discretized autonomous dynamical systems (ODEs) with polynomial nonlinearities exactly, learning neural structure from data generated by polynomial dynamics is tantamount to recovering the polynomial equations (terms and coefficients). These are fully interpretable, with obvious generalization and extrapolation abilities. And, the available structural basis is sound. We proceed with a structure learning experiment in this restricted setting though we emphasize that our approach is not limited in application {\it per se}. 

Here, consider the problem of learning neural structure and parameters from data generated by numerical solutions to the 
 chaotic Lorenz-63~\cite{Lorenz1963} system\footnote{Code may be found at github.com/sairavela/LorenzStructureLearn}, which is defined as:
   \begin{eqnarray}
       \label{eq:lorenz}
    \dot{x_1} &=& \sigma(x_2-x_1),\\
    \dot{x_2} &=& \rho x_1 - x_2 -x_1 x_3,\\
    \dot{x_3} &=& -\beta x_3 + x_1 x_2.
\end{eqnarray} 
 
 Suppose the starting model is a second-degree polynomial with nine terms per equation  
 \begin{equation}
 \mathbf{X} = \big(x_1, x_2, x_3, x_1x_2, x_1x_3, x_2x_3, x_1^2, x_2^2, x_3^2\big).
 \end{equation}
 There are thus $27$ unknown  parameters $\{a_{ij}\}$, where $i$ indexes $x_{1\dots 3}$ and $j$ indexes $\mathbf{X}$. The ``true" Lorenz equations are simulated from an arbitrary initial condition $\mathbf{x_0} = \left(-1.1, 2.2, -2.7\right)$ with parameters $\sigma = 10$, $\rho = 28$, and $\beta = 8/3$, and time step $dt=0.01$. The model equations are also simulated using a parameter ensemble of size $100$, each initialized i.i.d. from  Gaussian with mean $0$ and variance $100$. If the parameter matrix for the $k^\text{th}$ ensemble member is denoted $A_k: = \left[a_{ij}^k\right]$, then  $A_k\mathbf{X_t}$ are the instantaneous time derivatives. The system of The parameters are then updated using Equation \ref{enkf}.

Using the Ensemble Kalman learner with all $27$ possible parameters,   the model system converges to the right structure and parameter values in approximately $85$ iterations when we use an initial parameter mean $0$, variance $100$, and high-precision/small target variance of $1\times10^{-10}$. Actual parameters are recovered to within $3\times 10^{-4}$ with a posterior variance of $3.2\times 10^{-7}$ with the ``wrong" term coefficients going to zero. Convergence was repeatable and, remarkably, required no additional sparsity constraints.  

However, there are clear limitations. The initial model is arguably quite close to the true model because all true terms are given as options; in other words, the true model lies within the space of candidate models. In general, this is not the case, and the dimensionality of the starting model may be quite high.

\begin{figure}
    \centering
    \includegraphics[width=3.25in]{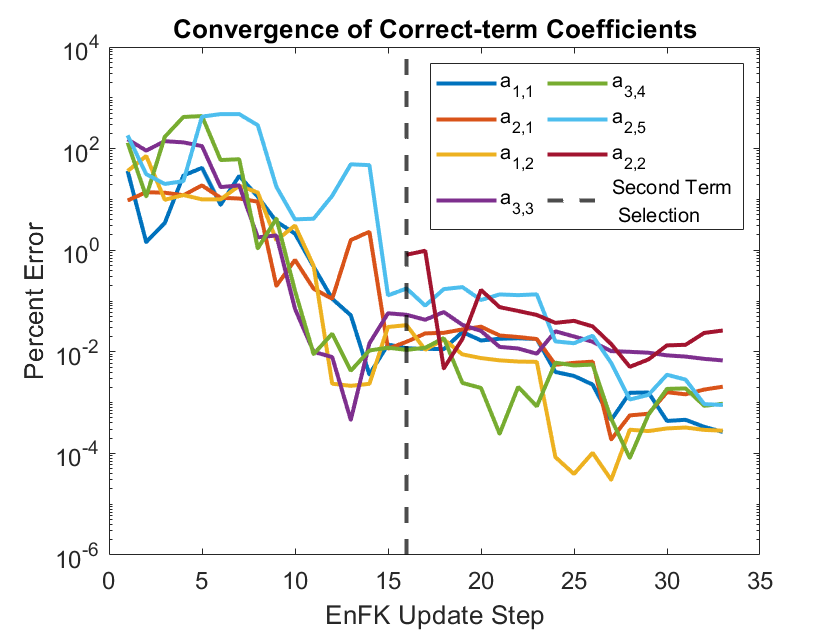}
    \includegraphics[width=3.25in]{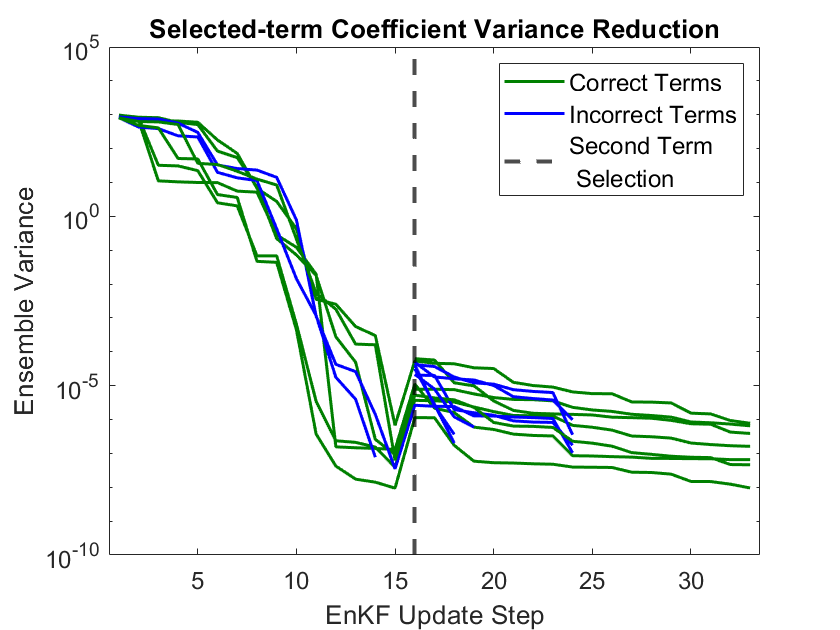}
    \caption{\it Convergence of the true parameters and variance reduction of chosen terms for learning the structure of the Lorenz-63 system.}
    \label{fig:lorconverge}
\end{figure}

The Informative Ensemble Kalman Learning approach, as discussed in the previous section, is the better option. Instead of automatically updating all terms using Ensemble Kalman Learning, we automatically select a small initial subset of terms as candidates and alternate between parameter estimation and term selection to sufficient prediction accuracy. To select terms, we first quantify the pairwise mutual information between each of the structure terms and each of the current model's three training error variables.  We calculate conditional pairwise mutual information assuming Gaussian ensembles, but non-Gaussian approaches are also applicable~\cite{tagade2014}. 
After that, we greedily select terms to maximize the cumulative sorted pairwise mutual information while minimizing the number of terms selected, a method akin to Akaike/Bayes selection criteria. The chosen terms adjoin the system equations, and Ensemble Kalman Learning proceeds for a specified variance reduction at the end of which terms with parameters values approaching zero leave the system equations. The selection cycle repeats. Variances are then rescaled and balanced in the new parameter ensemble, and Ensemble Learning proceeds. 

\begin{figure*}
    \centering
    \includegraphics[width=7in]{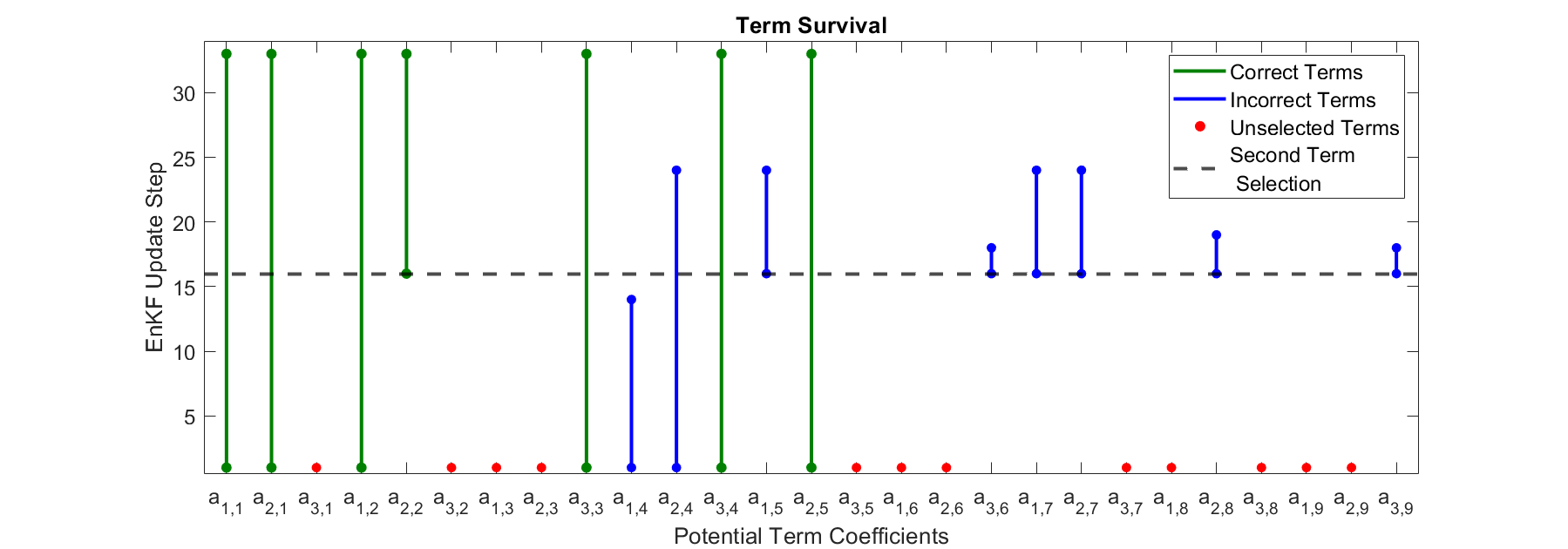}
    \caption{\it Survival of terms over iterations. Green are the correct terms, red are unselected, and blue are terms selected and later rejected.}
    \label{fig:lorsurvive}
\end{figure*}

\begin{figure*}
    \centering
    \includegraphics[width=7in]{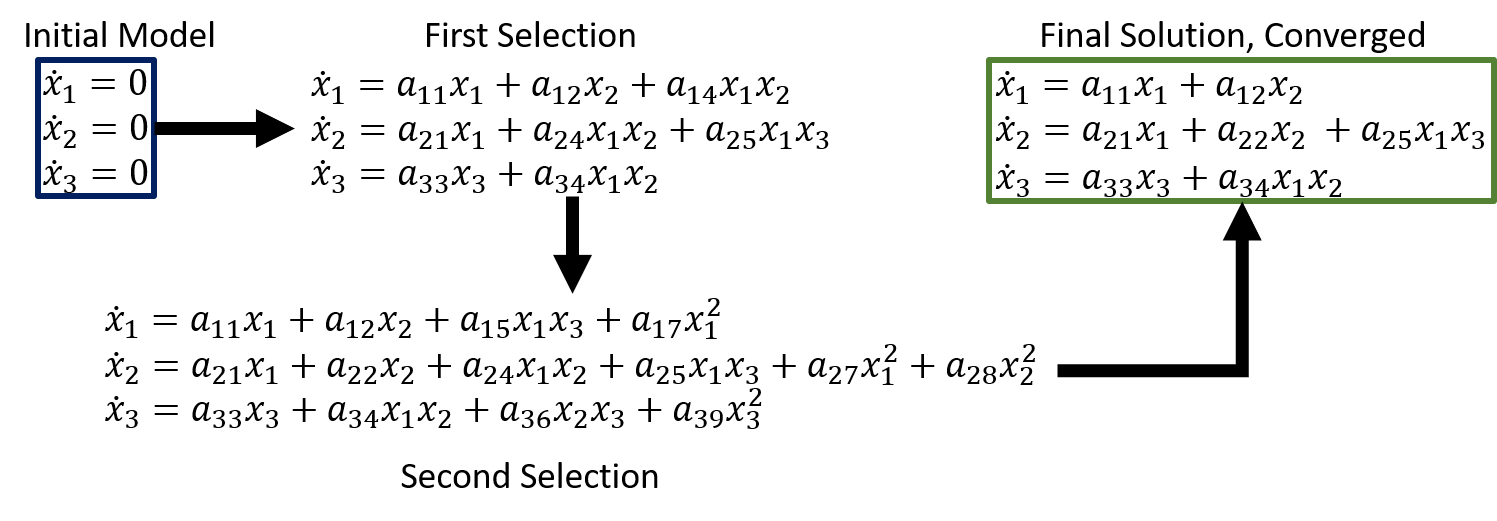}
    \caption{\it Ab Inito, iterations of Ensemble Kalman Learning follow a first selection and, after a second selection, convergence is achieved efficaciously in both terms and coefficients.}
    \label{fig:eqlearned}
\end{figure*}

By alternating the maximization of information gain with Ensemble Learning, we recover the Lorenz system equations from the initial model $\mathbf{\dot{x}_{1\dots3}} =0$ within approximately $35$ iterations requiring three or fewer selection steps. Not only is this more efficient but the incremental selection-rejection (prediction-correction) is automatic and overcomes the dimensionality concern. As shown in Figure~\ref{fig:lorconverge}, the true equations were recovered with parameter estimates within $1\%$. The final system structure learned is structurally exact:  $\dot{x}_1  = a_{11}x_1 + a_{12}x_2$, 
  $\dot{x}_2  = a_{21}x_1 + a_{22}x_2 + a_{25} x_1x_3$, and    $\dot{x}_3 = a_{33}x_3 + a_{34}x_1x_2$. The progression of term presence in the equations throughout the term selection process can be seen in Figure~\ref{fig:lorsurvive} and the equations themselves are shown in Figure~\ref{fig:eqlearned}.

    

\section{Conclusions}
\label{sec:concl}

 An informative optimization paradigm applies to neural learning.  Learning is a 2BVP and  the stochastic dynamics of Learning forms the basis of Informative Learning. Our Adaptive Ensemble Kalman Learning shows promising results on  two standard datasets, comparable to stochastic gradient descent. Other filters, smoothers, and controllers are also defined.   Informative Ensemble Kalman Learning maximizes information gain for Gaussian variables for variable selection. On structure learning, the correct network for the Lorenz system  equations are discovered {\em ab inito} faster and more tractably than a naive application. The special use of PolyNet NDS shows that the learned structure correspond exactly to the Lorenz equations.
 
 There are also a few limitations. Variable rate forgetting would be beneficial for adaptive learning. Although the applicability is general, the use of  low noise variance in the training data for Lorenz needs further experimentation. The sparsity considerations need additional work for smooth tractable sparsity control during learning. Finally, tractable non-Gaussian informative learning~\cite{tagade2014} may be needed. In future work, we will apply learn equations of natural hazards. 

\section*{Acknowledgements}
The authors thank the Earth Signals and Systems Group, MIT,  and anonymous reviewers for constructive reviews. Support from ONR grant N00014-19-1-2273 and VTSIX LTD are gratefully acknowledged.
%
%
%
%
\small
\bibliographystyle{splncs04}
\bibliography{refs}
\end{document}